\documentclass[a4paper,twoside]{article}
\pdfoutput=1

\usepackage{epsfig}
\usepackage{subcaption}
\usepackage{calc}
\usepackage{amssymb}
\usepackage{amstext}
\usepackage{amsmath}
\usepackage{amsthm}
\usepackage{multicol}
\usepackage{pslatex}
\usepackage{apalike}
\usepackage[linesnumbered]{algorithm2e}
\usepackage[bottom]{footmisc}

\usepackage[T1]{fontenc}

\usepackage[pdftex, 
            pdfauthor={Vincent A. Cicirello},
            pdftitle={Optimizing Genetic Algorithms Using the Binomial Distribution},
            pdfsubject={Cicirello, V.A. (2024). Optimizing Genetic Algorithms Using the Binomial Distribution.  In Proceedings of the 16th International Joint Conference on Computational Intelligence (IJCCI 2024), pages 159-169. https://doi.org/10.5220/0013038300003837},
            pdfkeywords={Bernoulli Trials, Binomial Random Variable, Genetic Algorithm, Mutation, Uniform Crossover}]{hyperref}

\usepackage{breakurl}

\usepackage{fancyhdr}

\fancypagestyle{specialfooter}{%
  \fancyhf{}
  
  \fancyfoot[L]{Accepted version. Cite published version as:\\Cicirello, V.A. (2024). Optimizing Genetic Algorithms Using the Binomial Distribution.  In {\it Proceedings of the 16th International Joint Conference on Computational Intelligence (IJCCI 2024)}, pages 159-169. \url{https://doi.org/10.5220/0013038300003837}}
}

\usepackage{SCITEPRESS}     % Please add other packages that you may need BEFORE the SCITEPRESS.sty package.

\SetKwProg{Fn}{function}{}{end}
\SetKwFunction{Rand}{Rand}
\SetKwFunction{FlipBit}{FlipBit}
\SetKwFunction{Sample}{Sample}
\SetKwFunction{Length}{Length}
\SetKwFunction{SimpleBitmask}{SimpleBitmask}
\SetKwFunction{OptimizedBitmask}{OptimizedBitmask}
\SetKwFunction{SimpleMutation}{SimpleMutation}
\SetKwFunction{OptimizedMutation}{OptimizedMutation}
\SetKwFunction{SimpleUniformCrossover}{SimpleCrossover}
\SetKwFunction{OptimizedUniformCrossover}{OptimizedCrossover}
\SetKwFunction{SimpleGeneration}{SimpleGeneration}
\SetKwFunction{OptimizedGeneration}{OptimizedGeneration}
\SetKwFunction{Crossover}{Crossover}

\begin{document}

\title{Optimizing Genetic Algorithms Using the Binomial Distribution}
\thispagestyle{specialfooter}

\author{\authorname{Vincent A. Cicirello \orcidAuthor{0000-0003-1072-8559}}
\affiliation{Computer Science, School of Business, Stockton University, 101 Vera King Farris Dr, Galloway, NJ, USA}
\email{vincent.cicirello@stockton.edu}
}


\abstract{Evolutionary algorithms rely very heavily on randomized behavior.
Execution speed, therefore, depends strongly on how we implement randomness,
such as our choice of pseudorandom number generator, or the algorithms 
used to map pseudorandom values to specific intervals or distributions. In
this paper, we observe that the standard bit-flip mutation of a genetic 
algorithm (GA), uniform crossover, and the GA control loop that determines 
which pairs of parents to cross are all in essence binomial experiments. 
We then show how to optimize each of these by utilizing a binomial distribution 
and sampling algorithms to dramatically speed the runtime of a GA relative to the 
common implementation. We implement our approach in the open-source Java library 
Chips-n-Salsa. Our experiments validate that the approach is orders of magnitude 
faster than the common GA implementation, yet produces solutions that are
statistically equivalent in solution quality.}

\onecolumn \maketitle \normalsize \setcounter{footnote}{0} \vfill

\section{\uppercase{Introduction}}
\label{sec:introduction}

The simulated evolutionary processes of genetic algorithms (GA), and other
evolutionary algorithms (EA), are stochastic and rely heavily on randomized 
behavior. Mutation in a GA involves randomly flipping bits. Cross points in 
single-point, two-point, and $k$-point crossover are chosen randomly.
Uniform crossover uses a process similar to mutation to randomly choose which
bits to exchange between parents. The decision of whether to cross a pair of 
parents during a generation is random. Selection operators usually randomize the 
selection of the population members that survive.

Random behavior pervades the EA. Thus, how randomness is implemented
significantly impacts performance. Many have explored the effects of the 
pseudorandom number generator (PRNG) on solution 
quality~\cite{Kromer2018,Rajashekharan2016,Kromer2013,Tirronen2011,Reese2009,Wiese2005,Wiese2005short,CantuPaz2002}, 
and others the effects of PRNG on execution times~\cite{FLAIRS2018,Nesmachnow2015}. 
Choice of PRNG isn't the only way to optimize an EA's random behavior. Our open source 
Java library Chips-n-Salsa~\cite{JOSS2020} optimizes randomness of its EAs and 
metaheuristics beyond the PRNG, such as in the choice of algorithms for Gaussian 
random numbers and random integers in an interval. Section~\ref{sec:related} further 
discusses related work.

This paper focuses on optimizing the randomness of bit-flip mutation, 
uniform crossover, and the determination of which pairs of parents 
to cross. Formally, all three of these are binomial experiments, with the 
relevant GA control parameters serving as the success probabilities for a 
sequence of Bernoulli trials. This leads to our approach (Section~\ref{sec:approach}) 
to each that replaces explicit iteration over bits or the population with a binomial 
distributed random variable and an efficient sampling algorithm. 

Others, such as Ye et al, previously observed that the number of bits mutated 
by the standard bit-flip mutation follows a binomial distribution~\cite{Ye2019}. 
However, examining the source code of their implementation~\cite{IOHalgorithm} 
reveals that they generate binomial random variates by explicitly iterating $n$ 
times for a bit-vector of length $n$, generating a total of $n$ uniform variates 
in the process. Their approach refactored where the explicit iteration occurs, 
but did not eliminate it. Ye et al were not trying to optimize the runtime of
the traditional GA. Rather, their reason for abstracting the mutated bit count
was to enable exploring alternative mutation operators, such as their normalized bit
mutation~\cite{Ye2019} in which the number of mutated bits follows a normal
distribution rather than a binomial.

In our approach, presented in detail in Section~\ref{sec:approach}, we eliminate
the $O(n)$ uniform random values generated during explicit iteration over bits. 
Instead, we use an efficient algorithm for generating binomial random 
variates~\cite{Kachitvichyanukul1988} that requires only $O(1)$ random numbers
on average. Additionally, we use a more efficient sampling algorithm for choosing
which bits to mutate. Ours is the first approach to use the binomial distribution
in this way to optimize the runtime of bit-flip mutation. Furthermore, we also 
apply the technique to optimize uniform crossover as well as the process of selecting 
which pairs of parents to cross.

Our experiments (Section~\ref{sec:experiments}) demonstrate the massive gains in 
execution speed that result from this approach. We will see that the optimized 
bit-flip mutation uses 70\%-99\% less time than the common implementation. The 
optimized uniform crossover uses 61\%-99\% less time than the common implementation. 
A GA using the optimized decision of which parents to cross, as well as the optimized 
operators, uses 78\%-97\% less time than the common implementation. We implemented the 
approach in the open source library Chips-n-Salsa~\cite{JOSS2020} and released 
the code of the experiments as open source to enable easily replicating the results. 
We wrap up with a discussion in Section~\ref{sec:conclusion}.

\section{\uppercase{Related Work}}\label{sec:related}

Several have shown that higher quality PRNGs that pass more rigorous 
randomness tests do not lead to better fitness, and that solution quality 
is insensitive to 
PRNG~\cite{Rajashekharan2016,Tirronen2011,Wiese2005,Wiese2005short,CantuPaz2002}. 
For example, the ablation study of Cant\'{u}-Paz shows that the PRNG used for 
selection, crossover, and mutation does not affect fitness, and that even using 
true random numbers does not improve fitness~\cite{CantuPaz2002}. 
Thus, choice of PRNG has little, if any, impact on solution fitness. 

Fewer have studied the impact of the PRNG on execution times, but the 
research that exists shows potential for massive efficiency gains. Nesmachnow 
et al analyzed the effects of several implementation factors 
on the runtime of EAs implemented in C~\cite{Nesmachnow2015}. They showed 50\%-60\%
improvement in execution times by using state-of-the-art PRNGs such 
as R250~\cite{Kirkpatrick1981} or Mersenne Twister~\cite{Matsumoto1998} rather 
than the C library's \verb|rand()| function. Cicirello showed that an adaptive 
permutation EA is 25\% faster~\cite{FLAIRS2018} when using the PRNG 
SplitMix~\cite{Steele2014} rather than the classic linear congruential~\cite{Knuth2}, 
and by implementing Gaussian mutation~\cite{Hinterding1995} with the ziggurat 
algorithm~\cite{Marsaglia2000,Leong2005} rather than the polar method~\cite{Knuth2}. 
There are also studies on the impact of PRNG and related algorithmic components on 
the runtime performance of randomized systems other than EAs, such as Metropolis 
algorithms~\cite{MaciasMedri2023}. Others explored the effects of programming 
language choice on GA runtime~\cite{MereloGuervos2017,MereloGuervos2016}.

The EAs of Chips-n-Salsa~\cite{JOSS2020} optimize randomness in ways beyond 
the choice of PRNG. For example, Chips-n-Salsa uses the algorithm of Lemire 
for random integers in an interval~\cite{Lemire2019} as implemented in the 
open source library $\rho\mu$~\cite{JOSS2022}, which is more than twice as fast as 
Java's built-in method due to a significantly faster approach to rejection 
sampling. Such bounded random integers are used by EAs in a variety of ways, 
such as the random cross sites for single-point, two-point, and $k$-point crossover.

\section{\uppercase{Approach}}\label{sec:approach}

We now present the details of our approach. In Section~\ref{sec:binomial},
we formalize the relationship between different processes within a GA and
binomial experiments. In Section~\ref{sec:sampling}, we describe our 
approach to random sampling. We derive our optimizations of random bitmask
generation, mutation, and uniform crossover in 
Sections~\ref{sec:bitmasks-opt},~\ref{sec:mutation-opt}, and~\ref{sec:crossover-opt},
respectively. We show how to optimize the process of determining which
pairs of parents to cross, as well as putting all of the optimizations together
in Section~\ref{sec:generation-opt}.

\begin{table}[t]
\caption{Notation shared across algorithm pseudocode.}\label{tab:notation}
\centering
\begin{tabular}{@{}lp{143pt}@{}}\hline
$B(n,p)$ & binomial distributed random integer \\\hline
\FlipBit{$v$, $i$} & flips bit $i$ of vector $v$ \\\hline
\Length{$v$} & simple accessor for length of $v$ \\\hline
\Rand{$a$, $b$} & random real in $[a, b)$ \\\hline
\Rand{$a$} & random integer in $[0,a)$ \\\hline
\Sample{$n$, $k$} & sample $k$ distinct integers from $\{0, 1, \ldots, n-1 \}$ \\\hline
$\oplus, \land, \lor, \lnot$ & bitwise XOR, AND, OR, and NOT \\\hline
\end{tabular}
\end{table}

Table~\ref{tab:notation} summarizes the notation and functions shared by the pseudocode of
the algorithms throughout this section. We assume that indexes into bit-vectors begin at 0.
The runtime of \FlipBit{$v$, $i$}, \Length{$v$}, and \Rand{$a$, $b$} is $O(1)$.
The runtime of the bitwise operators is $O(n)$ for vectors of length $n$. Our
bit-vector implementation utilizes an array of 32-bit integers, enabling exploiting
implicit parallelism, such as for bitwise operations (e.g., a bitwise operation
on two bit-vectors of length $n$ requires $\frac{n}{32}$ bitwise operations on 32-bit
integers).

\subsection{Binomial Variates and the GA}\label{sec:binomial}

A Bernoulli trial is an experiment with two possible outcomes, usually success 
and failure, and specified by success probability $p$. A binomial experiment is 
a sequence of $n$ independent Bernoulli trials with identical $p$~\cite{Larson1982}. 
The binomial distribution $B(n,p)$ is the discrete probability distribution of the 
number of successes in an $n$-trial binomial experiment with parameter $p$.

Our approach observes that GA mutation of a bit vector of length $n$ is a binomial 
experiment with $n$ trials, but the possible outcomes of each trial are ``flip-bit'' 
and ``keep-bit'' instead of success and failure. The number of bits flipped while 
mutating a bit vector of length $n$ must therefore follow binomial distribution 
$B(n,p_{m})$, where $p_{m}$ is the mutation rate. Uniform crossover of a pair of bit vectors 
of length $n$ is also a binomial experiment with $n$ trials, with the number of bits 
exchanged between parents following binomial distribution $B(n,p_{u})$, where $p_{u}$ is the 
probability of exchanging a bit between the parents. Choosing which pairs of parents 
to cross is likewise a binomial experiment with $n/2$ trials for population size $n$, 
and the number of crosses during a generation follows binomial distribution $B(n/2,p_{c})$ 
for crossover rate $p_{c}$.

We generate binomial random variates $B(n,p)$ with the BTPE 
algorithm~\cite{Kachitvichyanukul1988}, whose runtime is $O(1)$~\cite{ALG-24-007}.
This algorithm choice is critical to later analysis, as many alternatives 
do not have a constant runtime~\cite{Kuhl2017,Knuth2}.

\subsection{Sampling}\label{sec:sampling}

To sample $k$ distinct integers from the set $\{0, 1, \ldots, n-1 \}$, we
implement \Sample{$n$, $k$} using a combination of reservoir
sampling~\cite{Vitter1985}, pool sampling~\cite{Goodman1977}, and insertion 
sampling~\cite{cicirello2022applsci}, choosing the most efficient 
based on $k$ relative to $n$. 

\begin{algorithm}[t]
\caption{Random sampling.}\label{alg:sample}
\DontPrintSemicolon
\SetAlgoLined
\SetKwFunction{ReservoirSample}{ReservoirSample}%
\SetKwFunction{PoolSample}{PoolSample}%
\SetKwFunction{InsertionSample}{InsertionSample}%
\Fn{\Sample{$n$, $k$}}{
  \If{$k \geq \frac{n}{2}$}{
    \KwRet{\ReservoirSample{$n$, $k$}}
  }
  \If{$k \geq \sqrt{n}$}{
    \KwRet{\PoolSample{$n$, $k$}}
  }
  \KwRet{\InsertionSample{$n$, $k$}}
}
\end{algorithm}

\begin{algorithm}[t]
\caption{Random sampling component algorithms.}\label{alg:sample-helpers}
\DontPrintSemicolon
\SetAlgoLined
\SetKwFunction{ReservoirSample}{ReservoirSample}%
\SetKwFunction{PoolSample}{PoolSample}%
\SetKwFunction{InsertionSample}{InsertionSample}%
\Fn{\ReservoirSample{$n$, $k$}}{
  $s \leftarrow \text{an array of length}\ k$ \;
  \For{$i = 0$ \KwTo $k-1$}{
    $s[i] \leftarrow i$
  }
  \For{$i = k$ \KwTo $n-1$}{
    $j \leftarrow$ \Rand{$i+1$} \;
	\If{$j < k$}{
	  $s[j] \leftarrow i$
	}
  }
  \KwRet{$s$}
}
\;
\Fn{\PoolSample{$n$, $k$}}{
  $s \leftarrow \text{an array of length}\ k$ \;
  $t \leftarrow \text{an array of length}\ n$ \;
  \For{$i = 0$ \KwTo $n-1$}{
    $t[i] \leftarrow i$
  }
  $m \leftarrow n$ \;
  \For{$i = 0$ \KwTo $k-1$}{
    $j \leftarrow$ \Rand{$m$} \;
	$s[i] \leftarrow t[j]$ \;
	$m \leftarrow m - 1$ \;
	$t[j] \leftarrow t[m]$ \;
  }
  \KwRet{$s$}
}
\;
\Fn{\InsertionSample{$n$, $k$}}{
  $s \leftarrow \text{an array of length}\ k$ \;
  \For{$i = 0$ \KwTo $k-1$}{
    $v \leftarrow$ \Rand{$n-i$} \;
	$j \leftarrow k - i$ \;
	\While{$j < k$ \textbf{and} $v \geq s[j]$}{
	  $v \leftarrow v + 1$ \;
	  $s[j - 1] \leftarrow s[j]$ \;
	  $j \leftarrow j + 1$ \;
	}
	$s[j - 1] \leftarrow v$ \;
  }
  \KwRet{$s$}
}
\end{algorithm}

Algorithms~\ref{alg:sample} and~\ref{alg:sample-helpers} provide pseudocode of 
the details. See the original publications of the three sampling algorithms
for explanations for why each works. Our composition (Algorithm~\ref{alg:sample}) 
chooses from among the three sampling algorithms (Algorithm~\ref{alg:sample-helpers})
the one that requires the least random number generation
for the given $n$ and $k$. The runtime of reservoir sampling and pool 
sampling is $O(n)$, while the runtime of insertion sampling is $O(k^2)$. Pool 
sampling and insertion sampling each generate $k$ random integers, 
while reservoir sampling requires $(n-k)$ random integers. The runtime of the 
composite of these algorithms is $O(\min(n, k^2))$, and requires 
$\min(k, n-k)$ random integers~\cite{cicirello2022applsci}.
Minimizing random number generation is perhaps even more important
than runtime complexity, because although it is a constant time
operation, random number generation is costly.

\subsection{Optimizing Random Bitmasks}\label{sec:bitmasks-opt}

Our algorithms for mutation and crossover rely on random bitmasks of length 
$n$ for probability $p$ that a bit is a 1. Algorithm~\ref{alg:bitvector} 
compares pseudocode for the simple approach and the optimized 
binomial approach. The runtime (worst, average, and best cases) 
of the simple approach, \SimpleBitmask{$n$, $p$}, is $O(n)$. The 
runtime of the optimized version, \OptimizedBitmask{$n$, $p$}, 
is likewise $O(n)$, but the only $O(n)$ step initializes an 
all zero bit-vector in line 12. Most of the cost savings is due 
to requiring only $O(n\cdot\min(p, 1-p))$ random numbers on average 
(the call to \Sample{} on line 14), compared to \SimpleBitmask{$n$, $p$}, 
which always requires $n$ random numbers. When $p$ is small, such as
for mutation where $p$ is the mutation rate $p_{m}$, the cost 
savings from minimizing random number generation is especially 
advantageous as we will see in the experiments.

\begin{algorithm}[t]
\caption{Simple vs optimized bitmask creation.}\label{alg:bitvector}
\DontPrintSemicolon
\SetAlgoLined
\SetKwData{Indexes}{indexes}%
\Fn{\SimpleBitmask{$n$, $p$}}{
  $v \leftarrow \text{an all $0$ bit-vector of length}\ n$ \;
  \For{$i = 0$ \KwTo $n-1$}{
    \If{\Rand{$0.0$, $1.0$} $< p$}{
	  \FlipBit{$v$, $i$}
	}
  }
  \KwRet{$v$}
}
\;
\Fn{\OptimizedBitmask{$n$, $p$}}{
  $v \leftarrow \text{an all $0$ bit-vector of length}\ n$ \;
  $k \leftarrow B(n,p)$\;
  $\Indexes \leftarrow \Sample{n, k}$\;
  \For{$i \in \Indexes$}{
    \FlipBit{$v$, $i$}
  }
  \KwRet{$v$}
}
\end{algorithm}

Not shown in Algorithm~\ref{alg:bitvector} for presentation clarity,
our implementation of \OptimizedBitmask{$n$, $p$} treats $p=0.5$ as a 
special case by generating 32 random bits at a time with uniformly 
distributed random 32-bit integers. The \OptimizedBitmask{$n$, $p$}, 
as presented in Algorithm~\ref{alg:bitvector}, requires $\frac{n}{2}$ 
random bounded integers when $p=0.5$, while this special case 
treatment reduces this to $\frac{n}{32}$ random integers.

\subsection{Optimizing Mutation}\label{sec:mutation-opt}

Algorithm~\ref{alg:mutation} compares the common approach to 
mutation and our optimized approach. The $p_{m}$ is mutation rate. 
Since the simple approach, \SimpleMutation{$v$, $p_{m}$}, iterates over 
all $n$ bits, generating one random number for each, its runtime is $O(n)$. 
The runtime of \OptimizedMutation{$v$, $p_{m}$} is likewise $O(n)$ due to the 
$\oplus$ of vectors of length $n$, and the initialization of an 
all zero vector of length $n$. It is possible to eliminate these 
$O(n)$ steps with explicit iteration over only the mutated bits 
instead of using a bitmask. However, these steps are relatively 
inexpensive given the implicit parallelism associated with utilizing 
32-bit integers (e.g., bit operations on 32 bits at a time). The 
real time savings comes from reducing random number generation, 
where \OptimizedMutation{$v$, $p_{m}$} requires only $O(n\cdot\min(p_{m}, 1-p_{m}))$ 
random numbers on average via the call to \OptimizedBitmask{$n$, $p_{m}$} 
in line 12, while \SimpleMutation{$v$, $p_{m}$} requires $n$. Since the 
mutation rate $p_{m}$ is generally quite small, \OptimizedMutation{$v$, $p_{m}$} 
eliminates nearly all random number generation.

\begin{algorithm}[t]
\caption{Simple vs optimized mutation.}\label{alg:mutation}
\DontPrintSemicolon
\SetAlgoLined
\SetKwData{Mask}{bitmask}%
\Fn{\SimpleMutation{$v$, $p_{m}$}}{
  $n \leftarrow \Length{v}$ \;
  \For{$i = 0$ \KwTo $n-1$}{
    \If{\Rand{$0.0$, $1.0$} $< p_{m}$}{
	  \FlipBit{$v$, $i$}
	}
  }
}
\;
\Fn{\OptimizedMutation{$v$, $p_{m}$}}{
  $n \leftarrow \Length{v}$ \;
  $\Mask \leftarrow$ \OptimizedBitmask{$n$, $p_m$} \;
  $v \leftarrow v \oplus \Mask$
}
\end{algorithm}

\subsection{Optimizing Uniform Crossover}\label{sec:crossover-opt}

Algorithm~\ref{alg:crossover} compares the simple uniform crossover, 
determining the bits to exchange with iteration, versus our optimized 
version using our binomial trick. The $p_{u}$ is uniform crossover's 
parameter for the per-bit probability of bit exchange. Both versions 
are identical aside from how they generate a random bitmask. The runtime 
of both is $O(n)$ due to the bitwise operations. The time savings for 
\OptimizedUniformCrossover{$v_1$, $v_2$, $p_{u}$} derives from optimizing 
bitmask creation in line 11, leading the optimized uniform crossover to 
require $O(n \cdot \min(p_{u}, 1-p_{u}))$ random numbers, rather than the 
$O(n)$ random numbers required by the simple approach. For example, 
if $p_{u}=0.33$ or $p_{u}=0.67$, then the optimization requires 
only a third of the random numbers that would be needed if explicit bit 
iteration was used.

\begin{algorithm}[t]
\caption{Simple vs optimized uniform crossover.}\label{alg:crossover}
\DontPrintSemicolon
\SetAlgoLined
\SetKwData{Mask}{bitmask}%
\SetKwData{Temp}{temp}%
\Fn{\SimpleUniformCrossover{$v_1$, $v_2$, $p_{u}$}}{
  $n \leftarrow \Length{$v_1$}$ \;
  $\Mask \leftarrow$ \SimpleBitmask{$n$, $p_{u}$} \;
  $\Temp \leftarrow (v_1 \land \Mask) \lor (v_2 \land \lnot \Mask)$ \;
  $v_1 \leftarrow (v_2 \land \Mask) \lor (v_1 \land \lnot \Mask)$ \;
  $v_2 \leftarrow \Temp$
}
\;
\Fn{\OptimizedUniformCrossover{$v_1$, $v_2$, $p_{u}$}}{
  $n \leftarrow \Length{$v_1$}$ \;
  $\Mask \leftarrow$ \OptimizedBitmask{$n$, $p_{u}$} \;
  $\Temp \leftarrow (v_1 \land \Mask) \lor (v_2 \land \lnot \Mask)$ \;
  $v_1 \leftarrow (v_2 \land \Mask) \lor (v_1 \land \lnot \Mask)$ \;
  $v_2 \leftarrow \Temp$
}
\end{algorithm}

\subsection{Optimizing a Generation}\label{sec:generation-opt}

Algorithm~\ref{alg:generation} compares a simple implementation 
of a generation with an optimized version. The \SimpleGeneration{} 
explicitly iterates over the $\frac{n}{2}$ pairs of possible parents 
(lines 4--9), generating a uniform random variate for each to determine 
which parents produce offspring. The \OptimizedGeneration{} generates a single 
binomial random variate from $B(\lfloor\frac{n}{2}\rfloor, p_{c})$, where
$p_{c}$ is the crossover rate, to 
determine the number of crossover applications (line 18), and then iterates 
without need for additional random numbers (lines 19--21). 

\begin{algorithm}[t]
\caption{Simple vs optimized generation.}\label{alg:generation}
\DontPrintSemicolon
\SetAlgoLined
\SetKwData{Pop}{pop}%
\SetKwData{Pairs}{pairs}%
\SetKwFunction{Selection}{Selection}%
\Fn{\SimpleGeneration{\Pop, $p_{c}$, $p_{m}$}}{
  $n \leftarrow \Length{\Pop}$ \;
  $\Pop \leftarrow \Selection{\Pop}$\;
  \tcc{assume new population \Pop in random order}
  $\Pairs \leftarrow \lfloor\frac{n}{2}\rfloor$ \;
  \For{$i = 0$ \KwTo $\Pairs - 1$}{
    \If{\Rand{$0.0$, $1.0$} $< p_{c}$}{
	  \Crossover{$\Pop_i$, $\Pop_{i+\Pairs}$}
	}
  }
  \For{$i = 0$ \KwTo $n-1$}{
    \SimpleMutation{$\Pop_i$, $p_{m}$}
  }
}
\;
\Fn{\OptimizedGeneration{\Pop, $p_{c}$, $p_{m}$}}{
  $n \leftarrow \Length{\Pop}$ \;
  $\Pop \leftarrow \Selection{\Pop}$\;
  \tcc{assume new population \Pop in random order}
  $\Pairs \leftarrow B(\lfloor\frac{n}{2}\rfloor, p_{c})$\;
  \For{$i = 0$ \KwTo $\Pairs - 1$}{
    \Crossover{$\Pop_i$, $\Pop_{i+\Pairs}$}
  }
  \For{$i = 0$ \KwTo $n-1$}{
    \OptimizedMutation{$\Pop_i$, $p_{m}$}
  }
}
\end{algorithm}

Although the BTPE algorithm~\cite{Kachitvichyanukul1988} that we use to generate 
binomial random variates utilizes rejection sampling~\cite{Flury1990}, 
the average number of rejection sampling iterations is constant, 
and thus the average number of uniform variates needed by BTPE to 
generate a binomial is also constant~\cite{ALG-24-007}.
While \SimpleGeneration{} requires $O(\frac{n}{2})$ uniform random 
variates, \OptimizedGeneration{} requires only $O(1)$ random numbers. 

The \OptimizedMutation{} leads to additional 
speed advantage. The pseudocode uses a generic \Crossover{} operation 
rather than assuming any specific operator. In the experiments, we consider both 
uniform crossover, which we optimize, as well as 
single-point and two-point crossover, which don't have corresponding
binomial optimizations. The cross points of our single-point and two-point 
crossover operators are selected using a more efficient algorithm for
bounded random integers~\cite{Lemire2019} than Java's built-in method;
and the pair of indexes for two-point crossover are sampled using a
specially designed algorithm for small random samples~\cite{Cicirello2024spe}
that is significantly faster than general purpose sampling algorithms
such as those utilized earlier in Section~\ref{sec:sampling}. However, the
experiments in this paper that use single-point and two-point crossover
use these same optimizations for both the simple and optimized
experimental conditions.

\section{\uppercase{Experiments}}\label{sec:experiments}

We run the experiments on a Windows 10 PC with an AMD A10-5700, 
3.4 GHz processor and 8GB memory, and we use OpenJDK 64-Bit 
Server VM version 17.0.2. We implemented the optimized bit-flip
mutation, uniform crossover, and generation loop within
the open source library Chips-n-Salsa~\cite{JOSS2020}, 
and use version 7.0.0 in the experiments. All experiment source 
code is also open source. Table~\ref{tab:urls} provides the
relevant URLs.

\begin{table}[t]
\caption{URLs for Chips-n-Salsa and experiments.}\label{tab:urls}
\centering
\begin{tabular}{@{}lp{169pt}@{}}\hline
\multicolumn{2}{@{}l}{Chips-n-Salsa library} \\
Source & \url{https://github.com/cicirello/Chips-n-Salsa} \\
Website & \url{https://chips-n-salsa.cicirello.org/} \\
Maven & \url{https://central.sonatype.com/artifact/org.cicirello/chips-n-salsa} \\\hline
\multicolumn{2}{@{}l}{Experiments} \\
Source & \url{https://github.com/cicirello/optimize-ga-operators} \\\hline
\end{tabular}
\end{table}

The remainder of this section is organized as follows. Sections~\ref{sec:mutation-exp}
and~\ref{sec:crossover-exp} present our experiments with mutation and uniform crossover,
respectively. Then, in Section~\ref{sec:experiments-ga}, we provide experiments
comparing a fully optimized GA that optimizes the choice of which pairs of parents to
cross in addition to the mutation and uniform crossover optimizations.

\subsection{Mutation Experiments}\label{sec:mutation-exp} 

In our mutation experiments, we consider bit-vector length 
$n \in \{16, 32, 64, 128, 256, 512, 1024\}$,
and mutation rate $p_{m} \in \{\frac{1}{n}, \frac{2}{n}, \ldots, \frac{1}{4}\}$.
For each combination $(n, p_{m})$, we measure the CPU time to perform 100,000 mutations,
averaged across 100 trials. We test the significance of the differences between the
simple and optimized versions using Welch's unequal variances t-test~\cite{Welch1947,Derrick2016}.

\begin{figure*}[t!]
\centering
\subfloat[]{\epsfig{file = 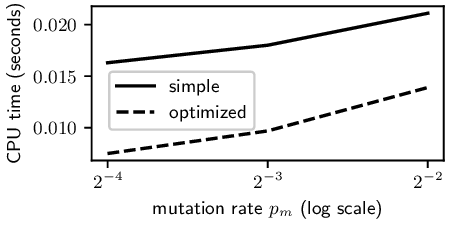, width = 2.75in}%
\label{fig:m16}}
\hfil
\subfloat[]{\epsfig{file = 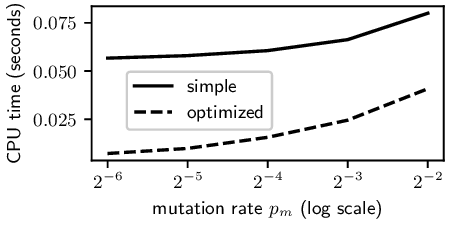, width = 2.75in}%
\label{fig:m64}} \\
\subfloat[]{\epsfig{file = 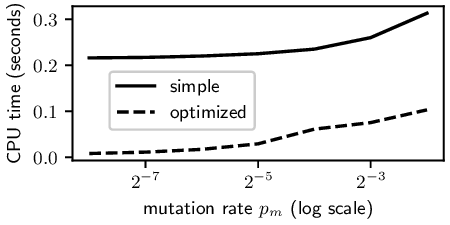, width = 2.75in}%
\label{fig:m256}}
\hfil
\subfloat[]{\epsfig{file = 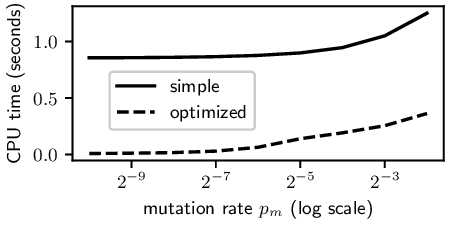, width = 2.75in}%
\label{fig:m1024}}
\caption{CPU time for $10^5$ mutations vs mutation rate $p_{m}$ for 
lengths: (a) 16 bits, (b) 64 bits, (c) 256 bits, (d) 1024 bits.}
\label{fig:mutation}
\end{figure*}

Figure~\ref{fig:mutation} visualizes $n \in \{16, 64, 256, 1024\}$.
Table~\ref{tab:mutation} summarizes results for $n=1024$. Data for all other cases 
is found in the GitHub repository, and is similar to the cases presented here.
 
\begin{table}[t]
\caption{CPU time for $10^5$ mutations for $n=1024$.}\label{tab:mutation}
\centering
\addtolength{\tabcolsep}{-0.26pt}
\begin{tabular}{@{}c|cc|cc@{}}\hline
    & \multicolumn{2}{c|}{CPU time (seconds)} & \% less & t-test \\
$p_{m}$ & simple & optimized & time & $p$-value \\\hline
$1/1024$ & 0.856 & 0.00906 & 98.9\% & $\sim 0.0$ \\
$1/512$ & 0.857 & 0.0119 & 98.6\% & $\sim 0.0$ \\
$1/256$ & 0.860 & 0.0173 & 98.0\% & $\sim 0.0$ \\
$1/128$ & 0.866 & 0.0300 & 96.5\% & $< 10^{-291}$ \\
$1/64$ & 0.877 & 0.0642 & 92.7\% & $\sim 0.0$ \\
$1/32$ & 0.899 & 0.140 & 84.5\% & $< 10^{-285}$ \\
$1/16$ & 0.946 & 0.192 & 79.7\% & $< 10^{-321}$ \\
$1/8$ & 1.05 & 0.255 & 75.7\% & $< 10^{-193}$ \\
$1/4$ & 1.25 & 0.363 & 71.0\% & $\sim 0.0$ \\
\hline
\end{tabular}
\addtolength{\tabcolsep}{0.26pt}
\end{table}

The optimized mutation leads to massive performance gains. For bit-vector 
length $n=1024$, the optimized mutation uses 71\% less time for high 
mutation rates. For typical low mutation rates, the optimized mutation 
uses 96\%--99\% less time than the simple implementation. All results are 
extremely statistically significant with t-test $p$-values very near zero.

\subsection{Uniform Crossover Experiments}\label{sec:crossover-exp}

We use the same bit-vector lengths
$n$ for the crossover experiments as we did for the mutation experiments;
and we consider $p_{u} \in \{0.1, 0.2, 0.3, 0.4, 0.5\}$ for the per-bit
probability of an exchange between parents. We do not consider $p_{u} > 0.5$
because any such $p_{u}$ has an equivalent counterpart $p_{u} < 0.5$ (e.g., 
exchanging 75\% of the bits between the parents leads to the same children
as if we instead exchanged the other 25\% of the bits). For each combination 
$(n, p_{u})$, we measure the CPU time to perform 100,000 uniform crossovers,
averaged across 100 trials, and again test significance with Welch's unequal 
variances t-test.

\begin{figure*}[t!]
\centering
\subfloat[]{\epsfig{file = 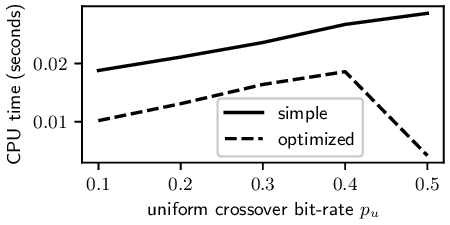, width = 2.75in}%
\label{fig:x16}}
\hfil
\subfloat[]{\epsfig{file = 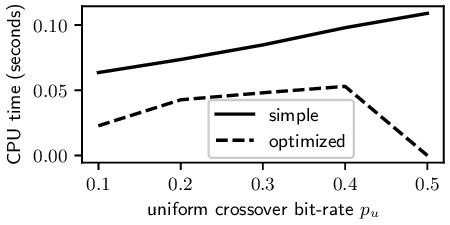, width = 2.75in}%
\label{fig:x64}} \\
\subfloat[]{\epsfig{file = 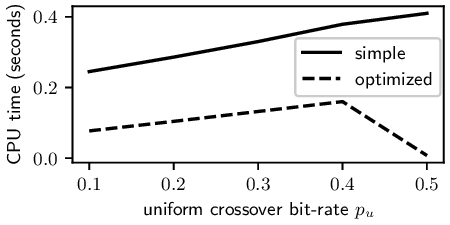, width = 2.75in}%
\label{fig:x256}}
\hfil
\subfloat[]{\epsfig{file = 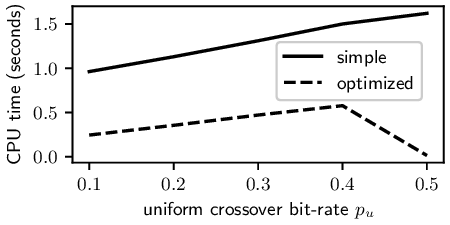, width = 2.75in}%
\label{fig:x1024}}
\caption{CPU time for $10^5$ uniform crosses vs uniform rate $p_{u}$ for 
lengths: (a) 16 bits, (b) 64 bits, (c) 256 bits, (d) 1024 bits.}
\label{fig:crossover}
\end{figure*}

Figure~\ref{fig:crossover} visualizes $n \in \{16, 64, 256, 1024\}$.
Table~\ref{tab:crossover} summarizes results for $n=1024$.
The data for all other cases is found in the GitHub repository, and is similar to the
cases presented here.

\begin{table}[t]
\caption{CPU time for $10^5$ uniform crosses for $n=1024$.}\label{tab:crossover}
\centering
\begin{tabular}{@{}c|cc|cc@{}}\hline
    & \multicolumn{2}{c|}{CPU time (seconds)} & \% less & t-test \\
$p_{u}$ & simple & optimized & time & $p$-value \\\hline
0.1 & 0.962 & 0.245 & 74.6\% & $\sim 0.0$ \\
0.2 & 1.13 & 0.355 & 68.5\% & $\sim 0.0$ \\
0.3 & 1.31 & 0.471 & 64.0\% & $< 10^{-308}$ \\
0.4 & 1.50 & 0.577 & 61.6\% & $< 10^{-312}$ \\
0.5 & 1.62 & 0.0142 & 99.1\% & $\sim 0.0$ \\
\hline
\end{tabular}
\end{table}

The optimized uniform crossover leads to similar performance gains
as seen with mutation. Specifically, for bit-vector length $n=1024$, 
the optimized uniform crossover uses approximately 60\% to 75\% less 
time than the simple implementation, except for the case of $p_{u}=0.5$ 
where the optimized version uses 99\% less time. The performance of 
$p_{u}=0.5$ is especially strong due to our special case treatment when 
generating random bitmasks (see earlier discussion in 
Section~\ref{sec:approach}). All results are extremely statistically 
significant with t-test $p$-values very near zero.

\subsection{GA Experiments}\label{sec:experiments-ga}

\begin{table*}[t!]
\caption{CPU time and average solution of 1024-bit OneMax using 1000 generations with 
uniform crossover ($p_{u}=0.33$).}\label{tab:ga33}
\centering
\begin{tabular}{@{}c||cc|cc||ccc@{}}\hline
& \multicolumn{2}{c|}{CPU time (seconds)} & \% less & t-test & \multicolumn{2}{c}{average solution} & t-test \\
$p_{c}$ & simple & optimized & time & $p$-value & simple & optimized & $p$-value \\\hline
0.05 & 0.869 & 0.0423 & 95.1\% & $\sim 0.0$    & 687.33 & 686.57 & 0.541\\
0.15 & 0.942 & 0.0698 & 92.6\% & $< 10^{-279}$ & 709.20 & 709.88 & 0.611\\
0.25 &  1.01 & 0.0950 & 90.6\% & $< 10^{-259}$ & 715.94 & 714.62 & 0.328\\
0.35 &  1.07 & 0.121 & 88.7\% & $\sim 0.0$     & 719.01 & 719.64 & 0.639\\
0.45 &  1.14 & 0.148 & 87.0\% & $< 10^{-314}$  & 720.77 & 721.57 & 0.512\\
0.55 &  1.21 & 0.174 & 85.6\% & $\sim 0.0$     & 721.54 & 722.01 & 0.718\\
0.65 &  1.27 & 0.200 & 84.3\% & $\sim 0.0$     & 722.97 & 724.94 & 0.158\\
0.75 &  1.34 & 0.224 & 83.3\% & $\sim 0.0$     & 723.89 & 722.16 & 0.163\\
0.85 &  1.41 & 0.252 & 82.1\% & $\sim 0.0$     & 725.60 & 726.18 & 0.655\\
0.95 &  1.47 & 0.277 & 81.2\% & $\sim 0.0$     & 724.62 & 724.73 & 0.929\\
\hline
\end{tabular}
\end{table*}

\begin{table*}[t!]
\caption{CPU time and average solution of 1024-bit OneMax using 1000 generations with 
uniform crossover ($p_{u}=0.49$).}\label{tab:ga}
\centering
\begin{tabular}{@{}c||cc|cc||ccc@{}}\hline
& \multicolumn{2}{c|}{CPU time (seconds)} & \% less & t-test & \multicolumn{2}{c}{average solution} & t-test \\
$p_{c}$ & simple & optimized & time & $p$-value & simple & optimized & $p$-value \\\hline
0.05 & 0.875 & 0.0467 & 94.7\% & $\sim 0.0$ & 690.05 & 689.17 & 0.454\\
0.15 & 0.956 & 0.0803 & 91.6\% & $< 10^{-312}$ & 711.34 & 712.19 & 0.572\\
0.25 &  1.03 & 0.114 & 88.9\% & $\sim 0.0$ & 717.99 & 717.08 & 0.439\\
0.35 &  1.11 & 0.148 & 86.7\% & $\sim 0.0$ & 720.71 & 722.44 & 0.162\\
0.45 &  1.19 & 0.182 & 84.8\% & $\sim 0.0$ & 722.24 & 720.16 & 0.102\\
0.55 &  1.27 & 0.216 & 83.0\% & $\sim 0.0$ & 723.10 & 723.38 & 0.826\\
0.65 &  1.35 & 0.250 & 81.5\% & $\sim 0.0$ & 723.67 & 724.12 & 0.707\\
0.75 &  1.43 & 0.283 & 80.2\% & $\sim 0.0$ & 725.33 & 724.64 & 0.582\\
0.85 &  1.51 & 0.318 & 79.0\% & $\sim 0.0$ & 725.65 & 724.22 & 0.272\\
0.95 &  1.59 & 0.352 & 77.9\% & $\sim 0.0$ & 724.85 & 724.44 & 0.761\\
\hline
\end{tabular}
\end{table*}

\begin{table*}[t!]
\caption{CPU time and average solution of 1024-bit OneMax using 1000 generations with 
single-point crossover.}\label{tab:ga1pt}
\centering
\begin{tabular}{@{}c||cc|cc||ccc@{}}\hline
& \multicolumn{2}{c|}{CPU time (seconds)} & \% less & t-test & \multicolumn{2}{c}{average solution} & t-test \\
$p_{c}$ & simple & optimized & time & $p$-value & simple & optimized & $p$-value \\\hline
0.05 & 0.836 & 0.0283 & 96.6\% & $\sim 0.0$ & 659.20 & 659.51 & 0.798 \\
0.15 & 0.838 & 0.0286 & 96.6\% & $\sim 0.0$ & 683.32 & 684.49 & 0.342 \\
0.25 & 0.839 & 0.0306 & 96.4\% & $< 10^{-260}$ & 693.77 & 694.79 & 0.413 \\
0.35 & 0.840 & 0.0308 & 96.3\% & $\sim 0.0$ & 701.48 & 703.06 & 0.230 \\
0.45 & 0.839 & 0.0325 & 96.1\% & $< 10^{-321}$ & 706.65 & 705.55 & 0.334 \\
0.55 & 0.842 & 0.0323 & 96.2\% & $\sim 0.0$ & 709.60 & 708.17 & 0.265 \\
0.65 & 0.843 & 0.0342 & 95.9\% & $\sim 0.0$ & 711.61 & 710.39 & 0.321 \\
0.75 & 0.843 & 0.0355 & 95.8\% & $< 10^{-310}$ & 713.20 & 713.91 & 0.571 \\
0.85 & 0.843 & 0.0366 & 95.7\% & $< 10^{-269}$ & 715.36 & 716.18 & 0.503 \\
0.95 & 0.850 & 0.0373 & 95.6\% & $< 10^{-223}$ & 716.60 & 716.36 & 0.854 \\
\hline
\end{tabular}
\end{table*}

\begin{table*}[t!]
\caption{CPU time and average solution of 1024-bit OneMax using 1000 generations with 
two-point crossover.}\label{tab:ga2pt}
\centering
\begin{tabular}{@{}c||cc|cc||ccc@{}}\hline
& \multicolumn{2}{c|}{CPU time (seconds)} & \% less & t-test & \multicolumn{2}{c}{average solution} & t-test \\
$p_{c}$ & simple & optimized & time & $p$-value & simple & optimized & $p$-value \\\hline
0.05 & 0.840 & 0.0291 & 96.5\% & $< 10^{-316}$ & 666.82 & 666.08 & 0.526 \\
0.15 & 0.840 & 0.0292 & 96.5\% & $\sim 0.0$      & 692.26 & 692.32 & 0.959 \\
0.25 & 0.839 & 0.0306 & 96.4\% & $< 10^{-296}$ & 702.23 & 703.29 & 0.394 \\
0.35 & 0.841 & 0.0313 & 96.3\% & $\sim 0.0$      & 708.16 & 709.99 & 0.124 \\
0.45 & 0.841 & 0.0333 & 96.0\% & $< 10^{-301}$ & 712.22 & 712.39 & 0.897 \\
0.55 & 0.842 & 0.0338 & 96.0\% & $\sim 0.0$      & 714.05 & 713.80 & 0.845 \\
0.65 & 0.844 & 0.0347 & 95.9\% & $< 10^{-311}$ & 718.86 & 715.63 & 0.00829 \\
0.75 & 0.845 & 0.0364 & 95.7\% & $\sim 0.0$      & 718.60 & 717.78 & 0.504 \\
0.85 & 0.845 & 0.0377 & 95.5\% & $< 10^{-318}$ & 719.54 & 718.86 & 0.611 \\
0.95 & 0.846 & 0.0383 & 95.5\% & $\sim 0.0$      & 720.30 & 721.55 & 0.320 \\
\hline
\end{tabular}
\end{table*}

Unlike mutation and crossover, it is not 
feasible to isolate the generation control logic of 
Algorithm~\ref{alg:generation} from the GA as it depends upon the genetic 
operators. Instead we experiment with a GA with simple 
vs optimized operators. We use the OneMax problem~\cite{Ackley1985} 
for its simplicity since our focus is on runtime performance of alternative 
implementations of operators, and use vector length $n=1024$ bits. 
All experimental conditions use population size 100, stochastic universal 
sampling~\cite{Baker1987} for selection, and mutation rate $p_{m}=\frac{1}{1024}$, 
which leads to an expected one mutated bit per population member per generation. We 
consider crossover rates $p_{c} \in \{0.05, 0.15, \ldots, 0.95\}$. Each GA run 
is 1000 generations, and we average results over 100 trials. We 
test significance with Welch's unequal variances t-test.

We consider four cases. Case (a) compares the simple versions of the 
generation logic, mutation, and uniform crossover ($p_{u}=0.33$) versus the 
optimized versions of these. Case (b) is similar, but with uniform crossover 
parameter $p_{u}=0.49$. We decided not to use $p_{u}=0.5$ to avoid the extra strong 
performance of our optimized bitmask generation for that special case. 
Cases (c) and (d) are as the others, but with single-point and two-point 
crossover, respectively. 

Table~\ref{tab:ga33} shows the results for case (a) uniform crossover
($p_{u}=0.33$). The optimized approach uses from approximately 81\% less time
for a high crossover rate ($p_{c}=0.95$) to 95\% less time for a low crossover 
rate ($p_{c}=0.05$). The runtime differences are extremely 
statistically significant with t-test $p$-values near 0.0.

Table~\ref{tab:ga} provides detailed results for case (b) uniform crossover
($p_{u}=0.49$). The results follow the same trend as the previous case. The approach 
optimized using the binomial distribution uses from 
approximately 78\% less time for a high crossover rate ($p_{c}=0.95$) to 95\% less 
time for a low crossover rate ($p_{c}=0.05$). The runtime differences are extremely 
statistically significant with t-test $p$-values near 0.0.

Table~\ref{tab:ga1pt} summarizes the results for case (c) single-point crossover.
Unlike uniform crossover, the speed difference does not vary by crossover rate.
Instead, the optimized approach uses approximately 95\% to 97\% less time than
the simple GA implementation for all crossover rates. All runtime differences
are extremely statistically significant with all t-test $p$-values very near 0.0.

\begin{figure*}[t]
\centering
\subfloat[]{\epsfig{file = 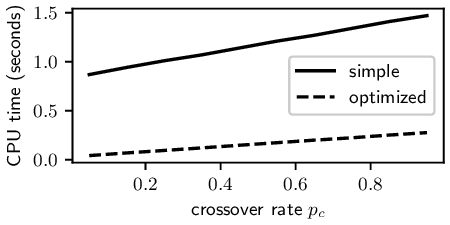, width = 2.75in}%
\label{fig:u33}}
\hfil
\subfloat[]{\epsfig{file = 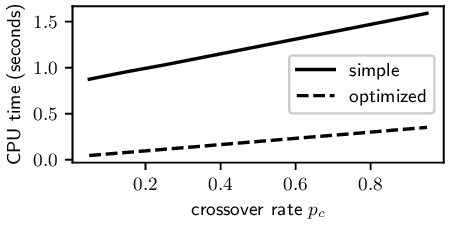, width = 2.75in}%
\label{fig:u49}} \\
\subfloat[]{\epsfig{file = 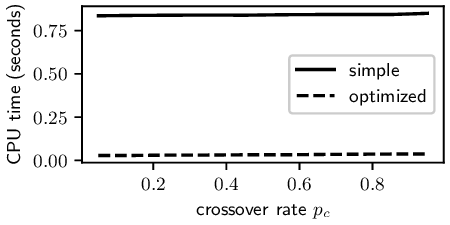, width = 2.75in}%
\label{fig:1p}}
\hfil
\subfloat[]{\epsfig{file = 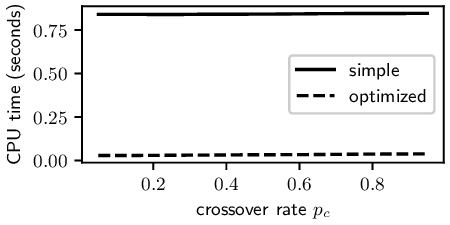, width = 2.75in}%
\label{fig:2p}}
\caption{CPU time for 1024-bit OneMax using 1000 generations with population size 100 
vs crossover rate $p_{c}$ for crossover operators: (a) uniform ($p_{u}=0.33$), 
(b) uniform ($p_{u}=0.49$), (c) single-point, (d) two-point.}
\label{fig:ga}
\end{figure*}

Table~\ref{tab:ga2pt} summarizes the results for case (d) two-point crossover.
The trend is the same as in the case of single-point crossover. The binomially 
optimized approach uses approximately 95\% to 97\% less time than the basic
implementation, and does not vary by crossover rate. All runtime differences
are extremely statistically significant with all t-test $p$-values very near 0.0.

Tables~\ref{tab:ga33},~\ref{tab:ga},~\ref{tab:ga1pt}, and~\ref{tab:ga2pt} also show 
average OneMax solutions (i.e., number of 1-bits) to demonstrate that the optimized 
approach does not statistically alter GA problem-solving behavior. For any given crossover 
operator and crossover rate $p_{c}$, there is no statistical significance in solution 
quality between the optimized and simple approach (i.e., high $p$-values). 

Figure~\ref{fig:ga} visualizes the results for all four cases. When uniform crossover
is used, runtime increases with crossover rate. The graphs appear to show constant runtime
when either single-point (Figure~\ref{fig:1p}) or two-point crossover (Figure~\ref{fig:2p}) 
is used. However, runtime is increasing in these cases, just very slowly as seen in 
the detailed results from Tables~\ref{tab:ga1pt} and~\ref{tab:ga2pt}.
The performance differential between the simple and optimized approaches doesn't vary by 
crossover rate when single-point (Figure~\ref{fig:1p}) or two-point crossover 
(Figure~\ref{fig:2p}) are used, with the optimized approach consistently using 95\% less time;
while the performance differential between the optimized and simple approaches does vary with
crossover rate when uniform crossover is used (Figures~\ref{fig:u33} and~\ref{fig:u49}).

\section{\uppercase{Conclusion}}\label{sec:conclusion}

In this paper, we demonstrated how we can significantly speed up the runtime of
some GA operators, including the common bit-flip mutation and uniform crossover,
by observing that such operators define binomial experiments (i.e., sequence of 
Bernoulli trials). This enables replacing explicit iteration that generates a 
random floating-point value for each bit to determine whether to flip (for mutation) 
or exchange (for crossover), with the generation of a single binomial random 
variate to determine the number of bits $k$, and an efficient sampling algorithm to 
choose the $k$ bits to mutate or cross. As a consequence, costly random number 
generation is significantly reduced. A similar approach is also seen for the 
generation logic that determines the number of parents to cross.

The technique is not limited to these operators, and is applicable for any 
operator that is controlled by some probability $p$ of including an element 
in the mutation or cross. For example, several evolutionary operators for 
permutations~\cite{cicirello2023ecta} operate in this way, including 
uniform order based crossover~\cite{Syswerda1991}, order 
crossover 2~\cite{Syswerda1991,Starkweather1991}, 
uniform partially matched crossover~\cite{cicirello2000gecco},
uniform scramble mutation~\cite{cicirello2023ecta},
and uniform precedence preservative crossover~\cite{Bierwirth1996}.
We adapt this approach in our implementations of all of these evolutionary
permutation operators in the open source library Chips-n-Salsa~\cite{JOSS2020}. 

\bibliographystyle{apalike}
{\small
\bibliography{ecta2024}}

\end{document}